\documentclass[10pt,twocolumn,letterpaper]{article}

\usepackage[pagenumbers]{cvpr}      
\definecolor{cvprblue}{rgb}{0.21,0.49,0.74}
\usepackage[pagebackref,breaklinks,colorlinks,allcolors=cvprblue]{hyperref}
\usepackage{multirow}

\def\paperID{*****} 
\def\confName{CVPR}
\def\confYear{2026}

\title{Pixel-to-4D: Camera-Controlled Image-to-Video Generation\\with Dynamic 3D Gaussians}

\author{
Melonie de Almeida
\and
Daniela Ivanova \and
Tong Shi \and
John H. Williamson \hspace{3em}
Paul Henderson \\[4pt]
University of Glasgow
}

\begin{document}
\maketitle
\begin{abstract}
Humans excel at forecasting the future dynamics of a scene given just a single image. Video generation models that can mimic this ability are an essential component for intelligent systems. Recent approaches have improved temporal coherence and 3D consistency in single-image-conditioned video generation. However, these methods often lack robust user controllability, such as modifying the camera path, limiting their applicability in real-world applications. Most existing camera-controlled image-to-video models struggle with accurately modeling camera motion, maintaining temporal consistency, and preserving geometric integrity. Leveraging explicit intermediate 3D representations offers a promising solution by enabling coherent video generation aligned with a given camera trajectory. Although these methods often use 3D point clouds to render scenes and introduce object motion in a later stage, this two-step process still falls short in achieving full temporal consistency, despite allowing precise control over camera movement.
We propose a novel framework that constructs a 3D Gaussian scene representation and samples plausible object motion, given a single image in a single forward pass. This enables fast, camera-guided video generation without the need for iterative denoising to inject object motion into render frames. Extensive experiments on the KITTI, Waymo, RealEstate10K, and DL3DV-10K datasets demonstrate that our method achieves state-of-the-art video quality and inference efficiency. Project page: \url{https://meloniedealmeida.github.io/Pixel-to-4D-Website/}
\end{abstract}    
\section{Introduction}
\label{sec:intro}

The human ability to predict dynamic changes in a scene from a single image is extraordinary; for example, we can easily estimate where the interacting cars and pedestrians in a street are likely to move in a second. 
Video Generation Models aim to mimic this human ability, and numerous approaches to this task have been developed over the past decade \cite{blattmann2023stable,xing2024dynamicrafter}.

Controllability in video generation enhances customization, realism, and usability. In particular, \textit{camera-controlled} video generation has gained significant attention recently \cite{zheng2024cami2v,he2024cameractrl,wang2023motionctrl,li2025realcam}. For example, CameraCtrl \cite{he2024cameractrl} is a text-to-video diffusion model conditioned on Plücker embedding of camera poses. Similarly, MotionCtrl \cite{wang2023motionctrl} integrates camera pose information into temporal transformers to improve control over motion. Meanwhile, CamI2V \cite{zheng2024cami2v}, CamCo~\cite{xu2024camco} and \cite{Tseng_2023_CVPR} incorporate epipolar attention for camera-conditioned video synthesis. 

However, these models still face challenges in accurately modeling camera motion, ensuring temporal consistency, preserving geometric integrity, and maintaining the style and lighting of the input image.
Leveraging explicit intermediate 3D representations in video prediction helps ensure temporal consistency and precise camera motion control. These representations enable frame rendering from varying viewpoints and support physically meaningful updates, maintaining coherent scene geometry and object motion over time.
For example, \cite{Lai_2021_ICCV} introduces a video auto-encoder that learns 3D scene geometry and camera motion, enabling 3D-aware video prediction from a single image. \cite{henderson2020unsupervised} model video using latent object appearance and pose variables rendered over static backgrounds. \cite{yarram2024forecasting} forecasts motion by generating point clouds from past frames, but it depends on depth estimation and inpainting, which can introduce errors.
Other approaches~\cite{li2025realcam,ren2025gen3c} estimate point clouds via monocular depth to capture camera motion, then inject object dynamics using video diffusion models. However, these methods can suffer from temporal inconsistencies and object-motion incoherence, and the sparsity of point clouds degrades video quality.

In order to perform video prediction with a latent 3D representation, we must choose a 3D representation that is realistic, fast to render, and easy to predict from a single frame. 3D Gaussian Splatting (3DGS) \cite{Kerbl2023-tu,zwicker2001ewa} is the state-of-the-art method for 3D reconstruction from 2D images. Unlike traditional point-cloud methods, 3DGS can fill gaps between points by adaptively scaling the size of the Gaussians. Recent works have extended 3DGS to 3D reconstruction given a single image, learnt from a dataset of multi-view images \cite{Szymanowicz2023-lx,szymanowicz2024flash3d}. These models predict pixel-aligned 3D Gaussians using standard image-to-image networks.

While latent 3D scene representations enable creating videos of static scenes, they cannot capture real-world scenes with dynamic objects.
We therefore go a step further and adapt the pixel-aligned Gaussian representation to support dynamic scenes by endowing each splat with 3D linear velocities and accelerations, and angular velocities and accelerations around the centroid of the object to which it belongs.
In contrast, most existing image-to-4D methods are limited to single-object scenes \cite{zheng2024unified,Lin2024-uh}, while approaches targeting real-world scenes rely on inpainting \cite{shen2023make} or video diffusion models \cite{sun2024dimensionx}, combined with depth estimation to reconstruct 3D geometry, often struggling to preserve fine details from the input image.

We introduce Pixel-to-4D, a framework that generates a 4D representation from a single image of a real-world scene in a single forward pass.
it uses this 4D representation to render predicted future frames along a user-controllable camera trajectory.
Our key contributions are:

\begin{itemize}
    \item We present a 4D representation for large-scale dynamic urban scenes with multiple layers of pixel-aligned static and dynamic Gaussian parameters. 
    \item We propose an efficient feedforward architecture that can generate this 4D Gaussian representation from a single image in a single pass, incorporating latent variables to capture uncertainty in future motion.
    \item We analyze the benefit of fusing information from a model (DINOv2~\cite{dino2023}) trained on large-scale datasets, showing that this improves image-to-4D and camera-controlled image-to-video prediction tasks. 
    
\end{itemize}

We perform a comprehensive evaluation of our method on the KITTI~\cite{geiger2013vision}, Waymo Open~\cite{mei2022waymo}, RealEstate10K~\cite{Zhou2018StereoMagnification}, and DL3DV-10K~\cite{Ling_2024_CVPR} datasets. The results show the effectiveness of our intermediate 4D representation for camera-controlled image-to-video generation. Our method surpasses baseline camera-controlled single-image-to-video models in PSNR, LPIPS, SSIM, and FVD while achieving lower inference time. 
This demonstrates that our method produces realistic predictions of the future, accurately following the user-specified camera motion and generating natural dynamics for moving objects as well as consistent depth renderings.

\begin{figure*}[t]
    \centering
    \includegraphics[width=\textwidth]{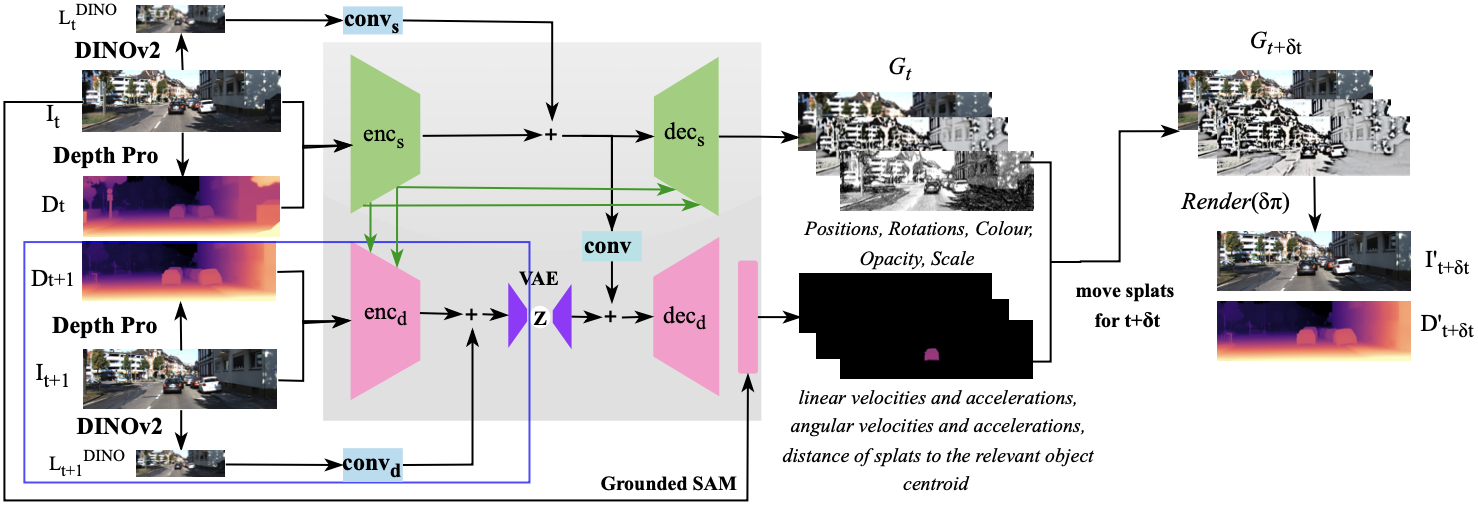}

    \vspace{-8pt}
    \caption{
    \textbf{Pixel-to-4D:} Given an input image \(I_t\), 
    \(\mathrm{enc}_s\) encodes \(I_t\) and its estimated depths \(D_t\) and fuses features from DINOv2.
    The combined features are decoded by \(\mathrm{dec_s}\) to predict static Gaussian parameters \(d,\Delta,r,s,\sigma,c\).
    Conditioned on the combined features, splat velocities \(v\) and accelerations \(a\) are generated using \(\mathrm{dec_{vae}}\) and \(\mathrm{dec_d}\) from latent Gaussian noise. These are aggregated over object segmentations to give final linear and angular velocities and accelerations.
    Then, a set of Gaussians \(G_{t+ \delta t}\) is derived from the static Gaussian parameters and velocities and accelerations, for a future time \(t+ \delta t\), from which the future frame is rendered with relative camera pose \(\delta\pi\). 
    The model is supervised by ground-truth future frames and their estimated depth-maps. During training, the inputs and model components within the blue box, \(I_{t+1}\), \(D_{t+1}\), \(L_{t+1}\), \(\mathrm{enc_d}\),\(\mathrm{conv_d}\) and \(\mathrm{enc_{vae}}\), are used to reconstruct $z$, and the model is optimized to align $z \sim \mathcal{N}(0, I)$. The green arrows represent skip connections from \(\mathrm{enc}_s\) to \(\mathrm{dec_s}\) and \(\mathrm{enc_d}\).}
    \label{fig:architecture}
\end{figure*}
\section{Related Work}
\label{sec:related_work}

\paragraph{Image-to-Video Generation and Camera-controllability.}
Recent advances in deep learning have led to significant progress in image-conditioned video prediction \cite{blattmann2023stable,xing2024dynamicrafter}. Camera-conditioned video prediction enables greater user control over the predicted video content. CameraCtrl \cite{he2024cameractrl} is an image-to-video diffusion model that conditions on Plücker embeddings of camera poses, while MotionCtrl~\cite{wang2023motionctrl} combines camera poses with temporal transformers. CamI2V \cite{zheng2024cami2v}, \cite{Tseng_2023_CVPR} integrate epipolar attention to improve camera motion accuracy. CVD 
\cite{kuang2024collaborative} and AC3D~\cite{Bahmani_2025_CVPR} jointly generate videos with different viewpoints showing consistent content.  These methods support limited camera trajectories.
The most similar methods in spirit to ours generate latent 3D representations and render these to ensure temporal and 3D consistency of the predicted video. \cite{henderson2020unsupervised} proposes an unconditional generative video model with latent object properties and appearances; however, this is limited to simple synthetic scenes. \cite{uvpf} proposes an unsupervised approach to predict future video frames from a single image by estimating the dynamic 3D structure of the scene.
\cite{Lai_2021_ICCV} introduces a video autoencoder that reconstructs videos via a latent 3D representation; however, it cannot handle object dynamics. \cite{yarram2024forecasting} generates a point cloud to predict motion and forecast videos. It depends on inpainting models for handling disocclusions and depth estimators for point-cloud generation, which may introduce inaccuracies.  
Recent methods generate a 3D point cloud from a single image, render videos along a camera trajectory, and inject object dynamics via video diffusion models \cite{li2025realcam,ren2025gen3c}. While effective for camera control, they suffer from key limitations: video diffusion often introduces temporal inconsistencies and weak alignment with the input image, point clouds can be sparse and inaccurate, and full-frame diffusion is computationally expensive. GEN3C \cite{ren2025gen3c} has shown good results only for supported trajectory patterns. In contrast, our method uses a 3D Gaussian representation that encodes object motion directly as linear and angular velocities and accelerations, ensuring coherent dynamics and temporal consistency. Moreover,  the explicit 4D representation enables efficient rendering of videos without frame-by-frame diffusion.


\paragraph{Novel view synthesis and image-to-3D Generation.}
Many studies have explored camera pose-conditioned novel view synthesis from a single input image via direct prediction of 2D pixels \cite{liu2023zero1to3zeroshotimage3d,liu2023one2345singleimage3d,long2024wonder3d,yu2023longtermphotometricconsistentnovel}. However, these methods struggle with inter-view consistency. 
Joint multi-view image generation partially mitigates this issue~\cite{liu2024syncdreamergeneratingmultiviewconsistentimages,shi2024mvdreammultiviewdiffusion3d,henderson2024sampling3dgaussianscenes}, but the lack of underlying geometric structures still results in flickering artifacts. Alternatively, some frameworks learn to predict radiance fields from a few images \cite{Anciukevicius2022-jf,Szymanowicz2023-md,anciukevičius2024denoisingdiffusionimagebasedrendering}; these remain limited to object-centric scenes.
A recent advance is to learn to predict 3D Gaussian parameters\cite{Kerbl2023-tu,zwicker2001ewa} per pixel using a deterministic 2D U-Net from a single image, enabling efficient image-to-3D reconstruction in a single forward pass \cite{Szymanowicz2023-lx,szymanowicz2024flash3d}. Some image-to-3D methods rely on pre-trained diffusion priors to optimize 3D representations using score distillation sampling (SDS) loss~\cite{tang2024dreamgaussiangenerativegaussiansplatting}. However, these methods are limited to static scenes.

\paragraph{Single image-to-4D Generation.}
Predicting 4D scenes from a single image has emerged as an active area of research. Building on image-to-3D methods, some image-to-4D approaches adopt a two-stage pipeline: first generating a static 3D representation~\cite{zheng2024unified,Lin2024-uh,shi2026splat}, followed by learning a deformation field to introduce temporal dynamics, typically guided by video diffusion models. However, these approaches are often constrained to isolated objects and orbital camera paths, limiting their scalability to complex, real-world environments.

Alternatively, Make-it-4d~\cite{shen2023make} inpaints RGB layers, estimates depths, constructs point clouds, and applies motion estimation. While effective in constrained settings, these methods struggle with forward-moving cameras due to compounded errors in depth and inpainting, which become evident under large viewpoint changes.
DreamDrive~\cite{mao2024dreamdrive} addresses this by generating reference frames using a video diffusion model, reconstructing into 3D using hybrid Gaussian representations. 
Inaccuracies in stereo depth estimation often cause Gaussian misalignments, resulting in visual artifacts in novel views.
DimensionX~\cite{sun2024dimensionx} uses separate controllable video diffusion models to handle spatial and temporal variations; this enables coarse controllability, but the method struggles to reproduce fine details, and it is limited to predefined camera motion trajectories.
\section{Methodology}

Our model predicts a 4D scene representation given a single image \(I_t\) (Fig.~\ref{fig:architecture}); this can be used to predict videos by rendering the scene at arbitrary small intervals \(\delta t\) into the future from user-specified camera viewpoints. Our method targets scenes with many dynamic objects and challenging camera motions, such as urban scenes.
Section \ref{sec:background} provides an overview of 3D Gaussian Splatting and its extension for single-image 3D reconstruction. Section~\ref{sec:4d_representation} introduces our proposed 4D representation. Section~\ref{sec:network} details our neural network model for 4D scene generation and video rendering. Finally, Section~\ref{sec:training} describes our training strategy.

\subsection{Background: 3D Gaussian Splatting} \label{sec:background}

Gaussian Splatting \cite{Kerbl2023-tu,zwicker2001ewa} is a point-based 3D representation that comprises $G$ colored Gaussians.
Each blob is a Gaussian density placed in 3D space, with location and shape determined by a mean $\mu_i \in \mathbb{R}^3$ and a covariance matrix $\Sigma_i \in \mathbb{R}^{3 \times 3}$. Splats also have a view-dependent color function $c_i(\nu) \in \mathbb{R}^3$ and an opacity $\sigma_i \in \mathbb{R}_+$.
An efficient differentiable function $\mathrm{render}(\mathcal{G}, \pi)$ is used to render an image $I$, where $\mathcal{G}$ is the set of Gaussians, and $\pi$ represents the camera viewpoint. 

To transform Gaussian Splatting into a learning framework, eliminating the need for test-time optimization and enabling reconstruction from a single image, Splatter-Image \cite{Szymanowicz2023-lx} uses an image-to-image U-Net \cite{RonnebergerFB15} trained on a dataset of multi-view images. Given an image $I \in \mathbb{R}^{3 \times H \times W}$, this model predicts an output \(\Phi(I) \in \mathbb{R}^{C \times H \times W}\), where the $C$ channels at each pixel represent the parameters of a Gaussian placed along the corresponding ray, with position defined by its depth along the ray.

\subsection{4D Scene Representation} \label{sec:4d_representation}

Our 4D scene representation models a dynamic scene from an input image using a set of Gaussian splats parameterized by pixel-aligned spatio-temporal parameters: 
\[
\begin{aligned}
\text{Initial depth:} & \quad d \in \mathbb{R}_+ \\
\text{Initial X-Y offsets:} & \quad \Delta \in \mathbb{R}^2 \\
\text{Initial X-Y-Z Velocities:} & \quad v \in \mathbb{R}^3 \\
\text{X-Y-Z Accelerations:} & \quad a \in \mathbb{R}^3 \\
\text{Initial Rotation:} & \quad r \in \mathbb{R}^4 \\
\text{Scale:} & \quad s \in \mathbb{R}^3 \text{ (log space)} \\
\text{Opacity:} & \quad \sigma \in \mathbb{R}_+;  
\sigma \in{(0,1]}  \\
\text{Color:} & \quad  
c \in{[-1,1]^3}\\
\end{aligned}
\]
with \(\Delta\) representing an offset in pixel-space, and \(d\), \(v\), and \(a\) defined in view-space with metric units.

Each pixel predicts parameters for $N \geq 1$ Gaussians: \(P = \{(\delta_i, \Delta_i, r_i, s_i, \sigma_i,c_i, \\ v_i, a_i)\}_{i=1}^N\). This representation can capture dis-occlusions due to dynamic objects. 

To predict accurate metric depths, we use pixel-wise depth estimates \(d_E\) from the state-of-the-art monocular predictor Depth-Pro \cite{bochkovskii2024depth}, combined with a predicted depth offset \(\delta_i\) for each Gaussian to correct inaccuracies in Gaussian depth estimation, cumulatively adjusting Gaussian positions by
\(
    d_i = d_E+\sum_{k=0}^{i} \delta_k
\).

The motion parameters of Gaussians are refined using an instance segmentation mask for each dynamic object, generated using \cite{ren2024grounded}, to ensure that Gaussians predicted from static regions of the input image are indeed static, while Gaussians predicted from the same object share the same linear and angular velocities and accelerations, modeling rigid body motion of each object. This works well for short intervals and regularizes against pathological solutions. 
Specifically, each object's linear velocity \(v^{lin}\) is the average over velocities $v$ of all its Gaussians, while its linear acceleration \(a^{lin}\) is the average over per-Gaussian accelerations $a$; its spin angular velocity \(\omega\) and spin angular acceleration \(\alpha\) are calculated by averaging those implied by its centroid and the per-Gaussian linear velocities and accelerations (the centroid is defined as the average 3D position of the Gaussians of the object).

The positions of the \(i\)\textsuperscript{th} Gaussian at the input frame's time \(t\) and a future time \(t+\delta t\) are: 
\begin{align}
    \mu_{i}(t) &= \left( \frac{(u^x+\Delta_i^x) d_i}{f_x}  , \frac{(u^y +\Delta_i^y)d_i}{f_y}, d_i  \right)
    \\
    \mu_{i}(t+\delta t) &= \mu_i(t)-p_i + v_i^{lin}\delta{t} + 0.5a_i^{lin}\delta{t}^2 \nonumber \\
&\quad +\omega^*_ip_i{\delta t} + 0.5\alpha^*_ip_i{\delta t}^2 \\
    r_i(t+\delta(t)) &=  \left(r_i(t) + \omega^*\delta{t} + 0.5 \alpha^*\delta{t}^2 \right).
\end{align}
where $p_i$ is the distance from the Gaussian to the centroid, $\omega^*_i$  and $\alpha^*_i$ are, respectively, the initial \emph{orbital} angular velocity and acceleration (i.e., rotational motion around the object centroid) for the Gaussian, and $f_x$ and $f_y$ are focal lengths.

We denote the differentiable Gaussian rasterization process \cite{Kerbl2023-tu} by \(\mathrm{render}(\mathcal{G}_{t+\delta t}, \delta \pi)\); this returns the rendered image \(I_{t+\delta t}\) and a corresponding depth map \(D_{t+\delta t}\), given the set of Gaussians \(\mathcal{G}_{t+\delta t}\) and the desired relative viewpoint \(\delta\pi\) for \(\delta t\).

\begin{figure*}[ht]
\centering
\renewcommand{\arraystretch}{0.05}
\setlength{\tabcolsep}{1pt}
\renewcommand{\arraystretch}{0.05}

\begin{tabular}{c c ccccccc c ccccccc}
& \multicolumn{1}{c}
{\includegraphics[height=1.1cm]{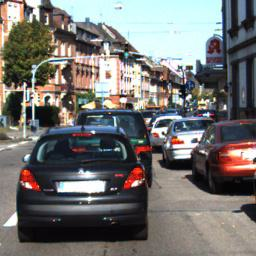}} &
  \multicolumn{13}{c}{\includegraphics[height=1.1cm]{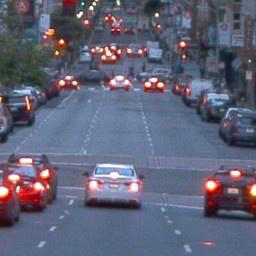}} \\
  \rotatebox{90}{~\scriptsize t+0.5}
& \includegraphics[height=1.1cm]{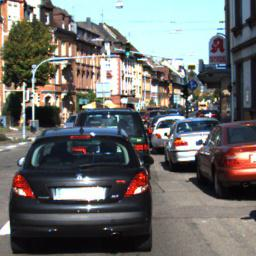} &
  \includegraphics[height=1.1cm]
  {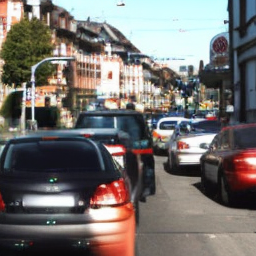} &
  \includegraphics[height=1.1cm]{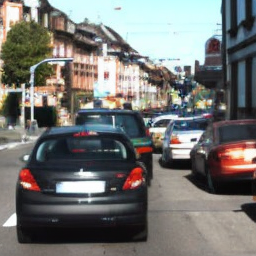} &
  \includegraphics[height=1.1cm]{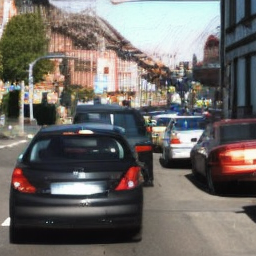} &
  \includegraphics[height=1.1cm]{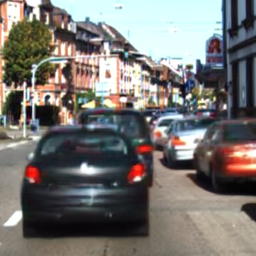} &
  \includegraphics[height=1.1cm]{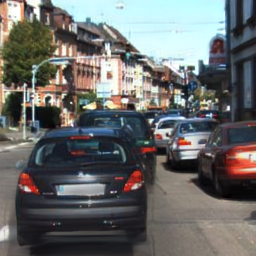} &
  \includegraphics[height=1.1cm]{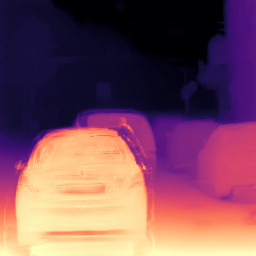} &
  \includegraphics[height=1.1cm]{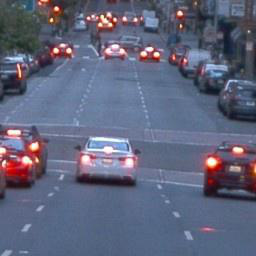} &
  \includegraphics[height=1.1cm]
  {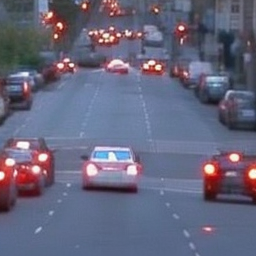} &
  \includegraphics[height=1.1cm]{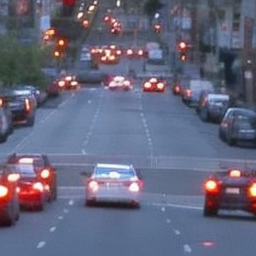} &
  \includegraphics[height=1.1cm]{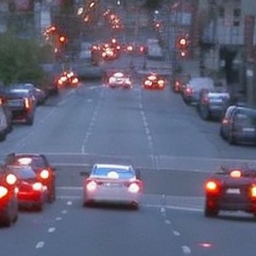} &
  \includegraphics[height=1.1cm]{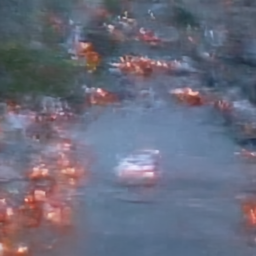} &
  \includegraphics[height=1.1cm]{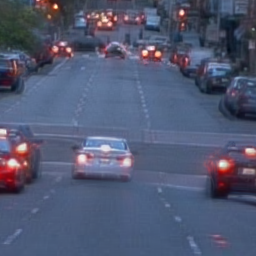} &
  \includegraphics[height=1.1cm]{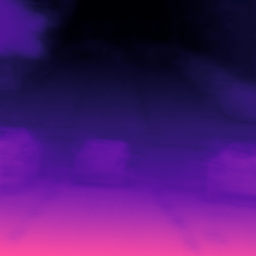}  \\
  \rotatebox{90}{~\scriptsize t+1.0}
& \includegraphics[height=1.1cm]{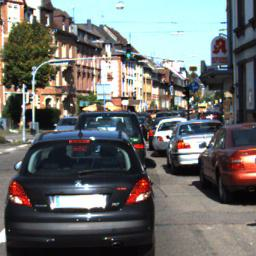} &
  \includegraphics[height=1.1cm]
  {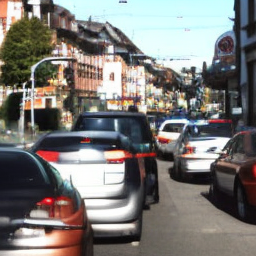} &
  \includegraphics[height=1.1cm]{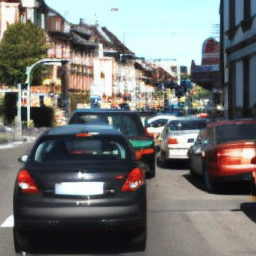} &
  \includegraphics[height=1.1cm]{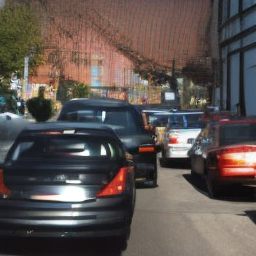} &
  \includegraphics[height=1.1cm]{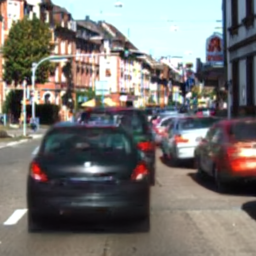} &
  \includegraphics[height=1.1cm]{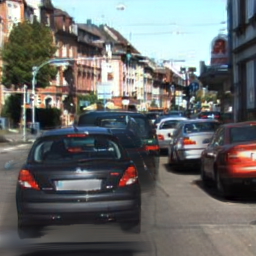} &
  \includegraphics[height=1.1cm]{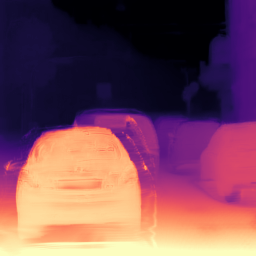} &
  \includegraphics[height=1.1cm]{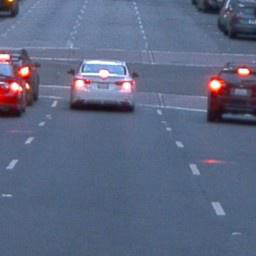} &
  \includegraphics[height=1.1cm]
  {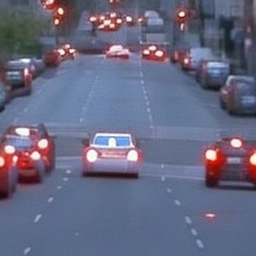} &
  \includegraphics[height=1.1cm]{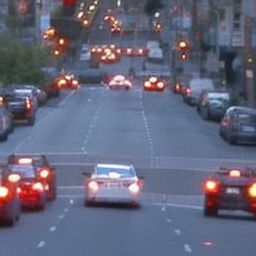} &
  \includegraphics[height=1.1cm]{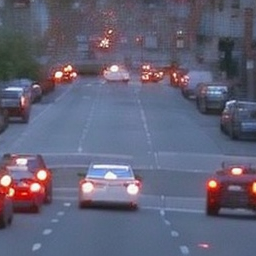} &
  \includegraphics[height=1.1cm]{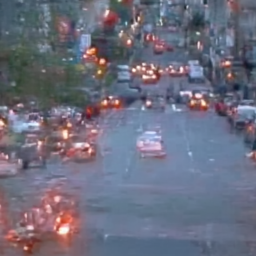} &
  \includegraphics[height=1.1cm]{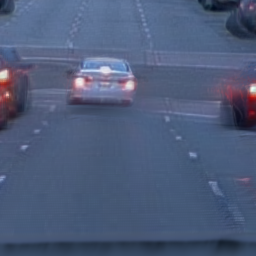} &
  \includegraphics[height=1.1cm]{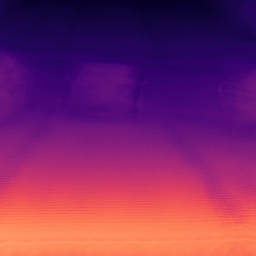}  \\
  \rotatebox{90}{~\scriptsize t+1.5}
& \includegraphics[height=1.1cm]{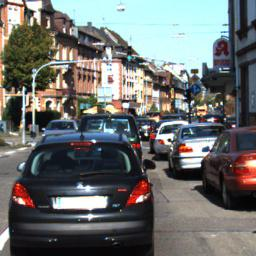} &
  \includegraphics[height=1.1cm]
  {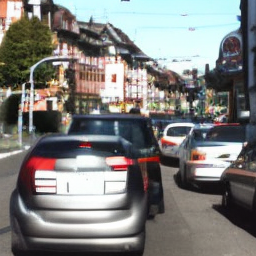} &
  \includegraphics[height=1.1cm]{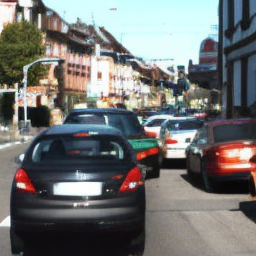} &
  \includegraphics[height=1.1cm]{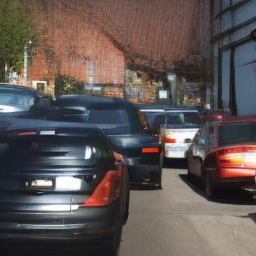} &
  \includegraphics[height=1.1cm]{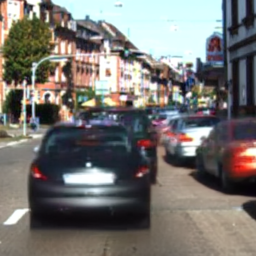} &
  \includegraphics[height=1.1cm]{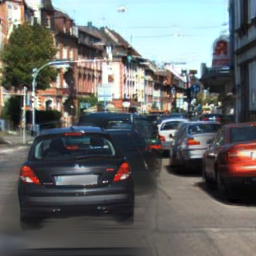} &
  \includegraphics[height=1.1cm]{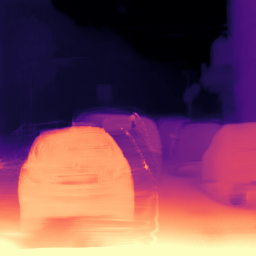} &
  \includegraphics[height=1.1cm]{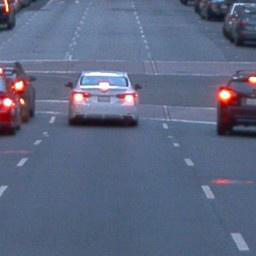} &
  \includegraphics[height=1.1cm]
  {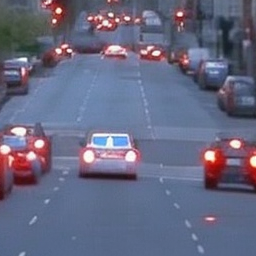} &
  \includegraphics[height=1.1cm]{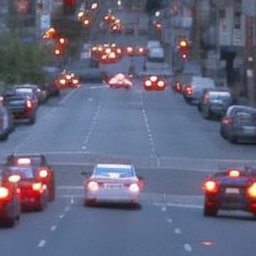} &
  \includegraphics[height=1.1cm]{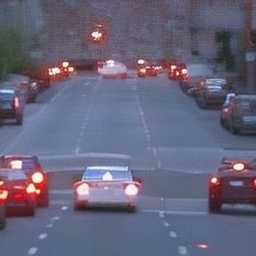} &
  \includegraphics[height=1.1cm]{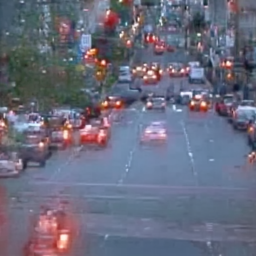} &
  \includegraphics[height=1.1cm]{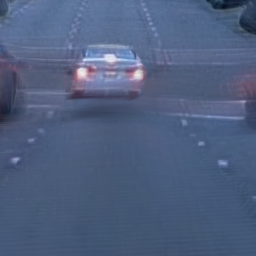} &
  \includegraphics[height=1.1cm]{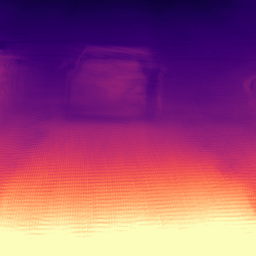}  \\ & \rotatebox{90}{\tiny GT}
& \rotatebox{90}{\tiny MotionCtrl} & \rotatebox{90}{\tiny CameraCtrl} & \rotatebox{90}{\tiny CamI2V} & 
  \rotatebox{90}{\tiny RealCam} & \rotatebox{90}{\tiny Ours} & \rotatebox{90}{\tiny Ours-depth} &
  \rotatebox{90}{\tiny GT}  & \rotatebox{90}{\tiny MotionCtrl} & \rotatebox{90}{\tiny CameraCtrl} & \rotatebox{90}{\tiny CamI2V} & 
  \rotatebox{90}{\tiny RealCam} & \rotatebox{90}{\tiny Ours} & \rotatebox{90}{\tiny Ours-depth} \\
& \multicolumn{7}{c}{\scriptsize KITTI} & \multicolumn{7}{c}{\scriptsize Waymo} \\
\end{tabular}

\begin{tabular}{c c cccccc  c ccccccc}

& \multicolumn{1}{c}{\includegraphics[height=1.1cm]{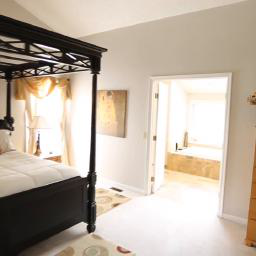}} &
  \multicolumn{13}{c}{\includegraphics[height=1.1cm]{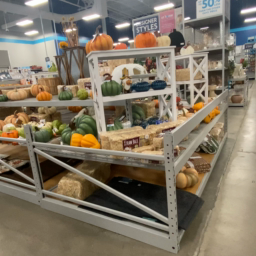}} \\
  \rotatebox{90}{~\scriptsize t+0.5}
& \includegraphics[height=1.1cm]{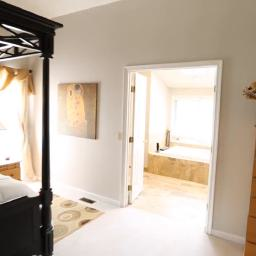} &
  \includegraphics[height=1.1cm]
  {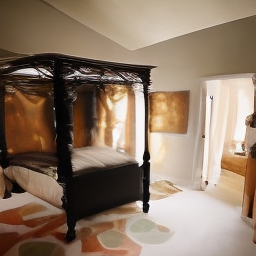} &
  \includegraphics[height=1.1cm]{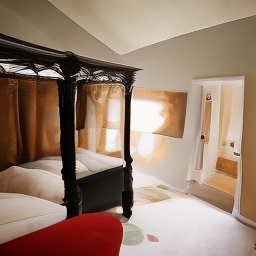} &
  \includegraphics[height=1.1cm]{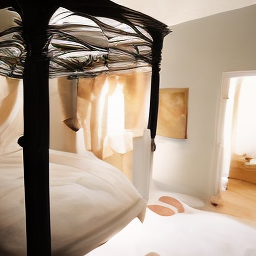} &
  \includegraphics[height=1.1cm]{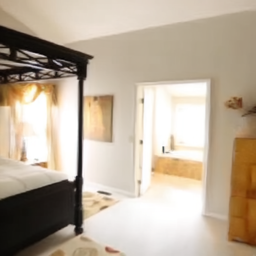} &
  \includegraphics[height=1.1cm]{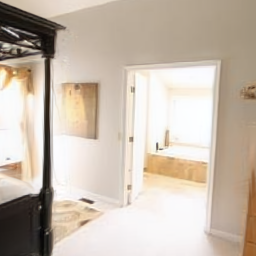} &
  \includegraphics[height=1.1cm]{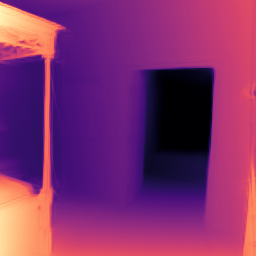} &
  \includegraphics[height=1.1cm]{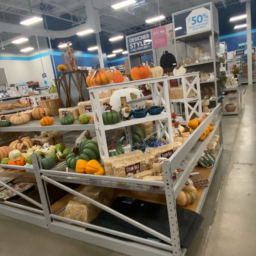} &
  \includegraphics[height=1.1cm]
  {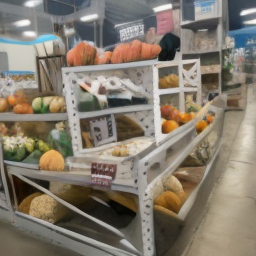} &
  \includegraphics[height=1.1cm]{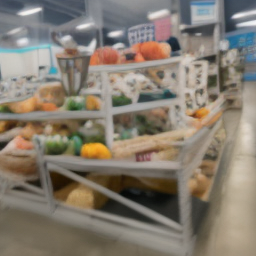} &
  \includegraphics[height=1.1cm]{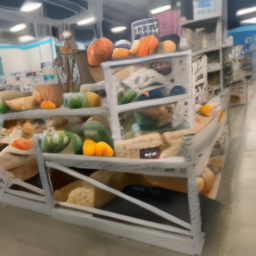} &
  \includegraphics[height=1.1cm]{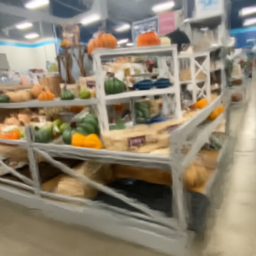} &
  \includegraphics[height=1.1cm]{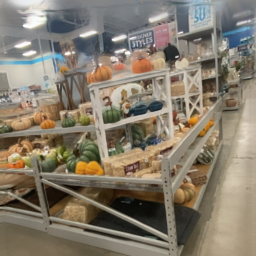} &
  \includegraphics[height=1.1cm]{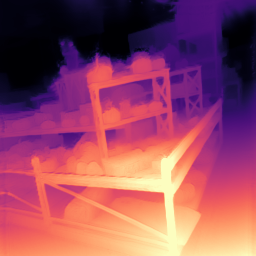}  \\
  \rotatebox{90}{~\scriptsize t+1.0}
& \includegraphics[height=1.1cm]{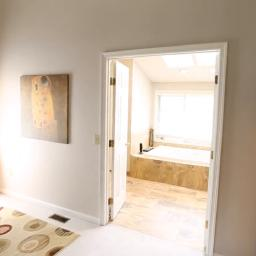} &
  \includegraphics[height=1.1cm]
  {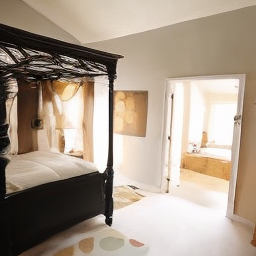} &
  \includegraphics[height=1.1cm]{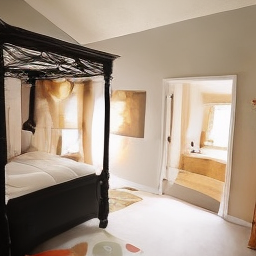} &
  \includegraphics[height=1.1cm]{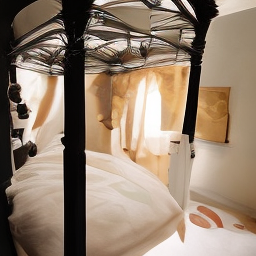} &
  \includegraphics[height=1.1cm]{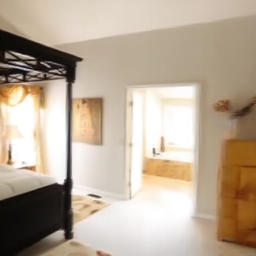} &
  \includegraphics[height=1.1cm]{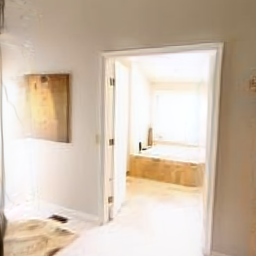} &
  \includegraphics[height=1.1cm]{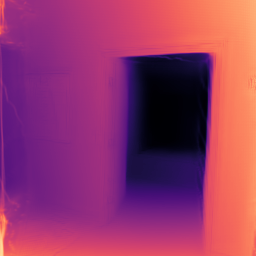} &
  \includegraphics[height=1.1cm]{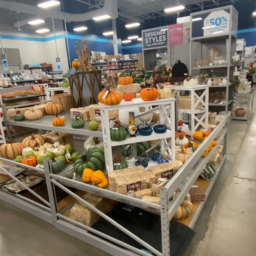} &
  \includegraphics[height=1.1cm]
  {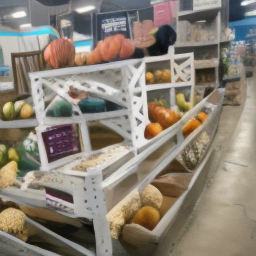} &
  \includegraphics[height=1.1cm]{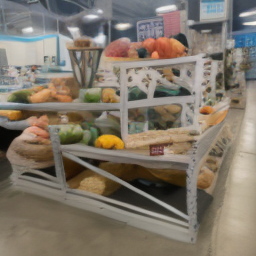} &
  \includegraphics[height=1.1cm]{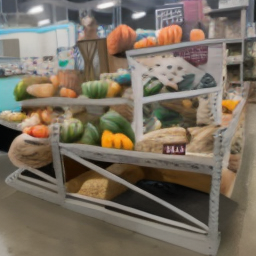} &
  \includegraphics[height=1.1cm]{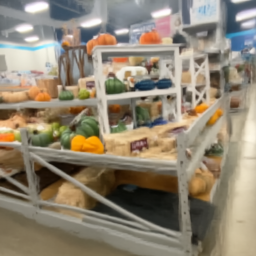} &
  \includegraphics[height=1.1cm]{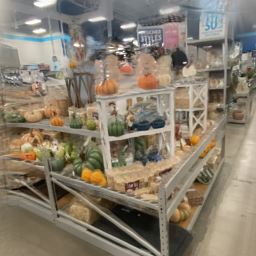} &
  \includegraphics[height=1.1cm]{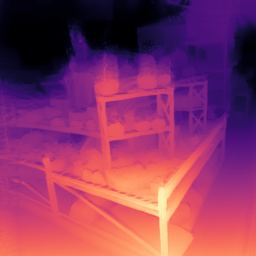}  \\
  \rotatebox{90}{~\scriptsize t+1.5}
& \includegraphics[height=1.1cm]{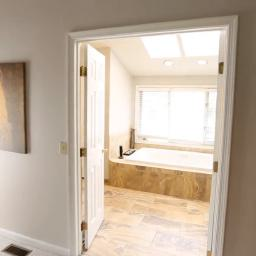} &
  \includegraphics[height=1.1cm]
  {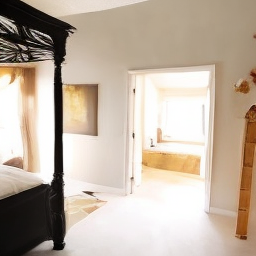} &
  \includegraphics[height=1.1cm]{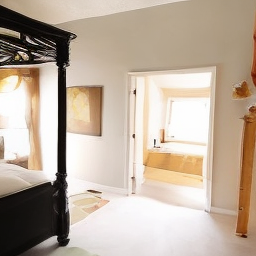} &
  \includegraphics[height=1.1cm]{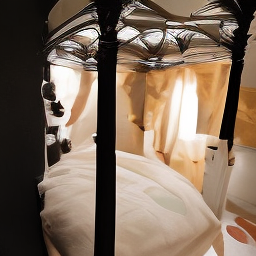} &
  \includegraphics[height=1.1cm]{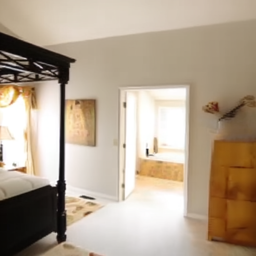} &
  \includegraphics[height=1.1cm]{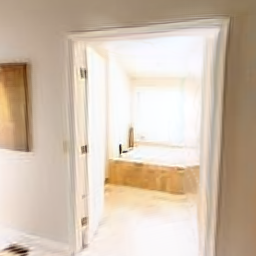} &
  \includegraphics[height=1.1cm]{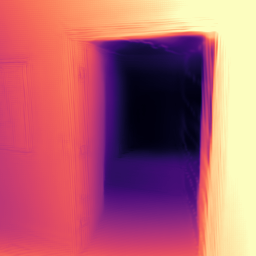} &
  \includegraphics[height=1.1cm]{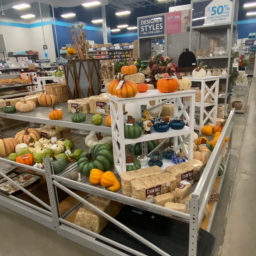} &
  \includegraphics[height=1.1cm]
  {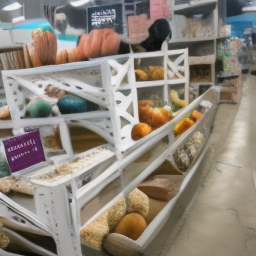} &
  \includegraphics[height=1.1cm]{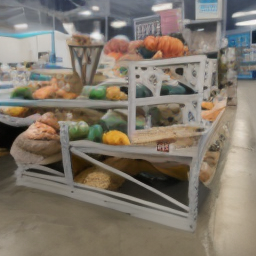} &
  \includegraphics[height=1.1cm]{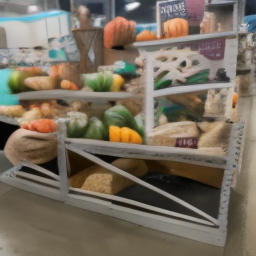} &
  \includegraphics[height=1.1cm]{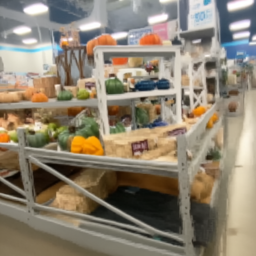} &
  \includegraphics[height=1.1cm]{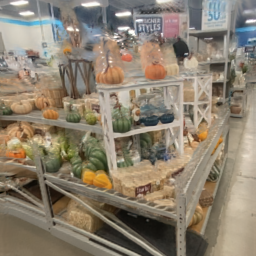} &
  \includegraphics[height=1.1cm]{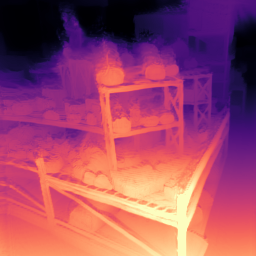}  \\
& \rotatebox{90}{\tiny GT} & \rotatebox{90}{\tiny MotionCtrl} & \rotatebox{90}{\tiny CameraCtrl} & \rotatebox{90}{\tiny CamI2V} & 
  \rotatebox{90}{\tiny RealCam} & \rotatebox{90}{\tiny Ours} & \rotatebox{90}{\tiny Ours-depth} &
  \rotatebox{90}{\tiny GT} & \rotatebox{90}{\tiny MotionCtrl} & \rotatebox{90}{\tiny CameraCtrl} & \rotatebox{90}{\tiny CamI2V} & 
  \rotatebox{90}{\tiny RealCam} & \rotatebox{90}{\tiny Ours } & \rotatebox{90}{\tiny Ours-depth} \\
& \multicolumn{7}{c}{\scriptsize RealEstate10K} & \multicolumn{7}{c}{\scriptsize DL3DV-10K} \\
\end{tabular}
\caption{Qualitative comparisons on four datasets. Each block shows the input frame at $t=0$ and ground truth and generated results at $t+0.5s$, $t+1s$, and $t+1.5s$ for five different methods. Our method consistently outperforms baseline approaches in both camera controllability and visual quality.}
\label{fig:qual_results}
\end{figure*}

\subsection{Predicting 4D Scenes from One Image} \label{sec:network}

We next describe how to predict the 4D scene representation from an input image. 
An encoder \(\mathrm{enc_s}\) processes the input frame \(I_t\) and its estimated depth-map \(D_t\) to produce an intermediate latent representation. 
To effectively leverage large-scale pretrained knowledge, we employ DINOv2~\cite{dino2023}, a self-supervised vision transformer, to extract features from $I_t$ for predicting static and dynamic Gaussian parameters. The DINOv2 output is passed through a convolution, $\mathrm{conv_s}$, to match the dimensionality of $\mathrm{enc_s}(I_t, D_t)$, and the results are summed to form the latent scene representation $L_s$
:
\begin{equation}
    L_s = \mathrm{enc_s}(I_t, D_t)+\mathrm{conv_s}(\text{DINOv2}(I_t))
\end{equation}
Then, the decoder \(\mathrm{dec_s}\)  predicts an output, \(G_t\), for static parameters \(\delta\), \(\Delta\), \(r\), \(s\), \(\sigma\), \(c\) for each Gaussian:
\begin{align}
    G_t &= \mathrm{dec_s}(L_s )
\end{align}
To facilitate faster learning of static Gaussian parameters such as color, which can be directly inferred from pixel values, \(\mathrm{dec_s}\) incorporates skip connections from \(\mathrm{enc_s}\).

The velocity of objects in the input image is uncertain; therefore, we sample plausible velocities conditioned on features of the input image, \(L_s\). To do so, we sample latent noise \(z\) from a standard diagonal Gaussian distribution and decode it using \(\mathrm{dec_{vae}}\) to match the dimensions of \(L_s\). The decoded noise is added to \(\mathrm{conv}(L_s)\) and passed to the dynamic decoder \(\mathrm{dec_d}\), which predicts the velocities, \(V=\{v\}\), and accelerations, \(A=\{a\}\), for all Gaussians jointly as follows:
\begin{align}
    z_{sample} &\sim \mathcal{N}(0, I) \\
    \quad V, A &= \mathrm{dec_d}\big(\mathrm{conv}(L_s) + \mathrm{dec_{vae}}(z_{sample})\big)
\end{align}

\begin{table}[t]
    \centering
    \small
    \setlength{\tabcolsep}{2.5pt}
    \renewcommand{\arraystretch}{0.75}
    \resizebox{\linewidth}{!}{
    \begin{tabular}{@{}l l c c c c c c@{}}
        \toprule
        Dataset & Model 
        & PSNR $\uparrow$ 
        & LPIPS $\downarrow$ 
        & SSIM $\uparrow$ 
        & FVD $\downarrow$ 
        & Depth $\downarrow$ 
        & Time (s) $\downarrow$ \\
        \midrule
        \multirow{5}{*}{KITTI}
        & MotionCtrl & 10.6 & 0.507 & 0.146 & 50.8 & -- & 11.8 \\
        & CameraCtrl & 11.2 & 0.490 & 0.220 & 75.7 & -- & 12.5 \\
        & CamI2V     & 11.9 & 0.488 & 0.199 & 70.5 & -- & 17.5 \\
        & RealCam    & 13.6 & 0.447 & 0.301 & 58.5 & -- & 9.8  \\
        & Ours       & \textbf{15.2} & \textbf{0.387} & \textbf{0.368} & \textbf{33.8} & \textbf{0.268} & \textbf{5.9} \\
        \midrule
        \multirow{5}{*}{Waymo}
        & MotionCtrl & 14.6 & 0.491 & 0.280 & 43.3 & -- & -- \\
        & CameraCtrl & 15.3 & 0.501 & 0.330 & 82.2 & -- & -- \\
        & CamI2V     & 15.4 & 0.499 & 0.328 & 50.3 & -- & -- \\
        & RealCam    & 17.0 & 0.449 & 0.424 & 47.8 & -- & -- \\
        & Ours       & \textbf{19.4} & \textbf{0.352} & \textbf{0.553} & \textbf{30.9} & \textbf{0.343} & -- \\
        \midrule
        \multirow{5}{*}{RE10K}
        & MotionCtrl & 12.1 & 0.554 & 0.201 & 57.5 & -- & -- \\
        & CameraCtrl & 12.7 & 0.529 & 0.233 & 62.8 & -- & -- \\
        & CamI2V     & 12.6 & 0.538 & 0.231 & 56.7 & -- & -- \\
        & RealCam    & 14.0 & 0.483 & 0.312 & 49.1 & -- & -- \\
        & Ours       & \textbf{18.2} & \textbf{0.314} & \textbf{0.495} & \textbf{24.2} & \textbf{0.195} & -- \\
        \midrule
        \multirow{5}{*}{DL3DV}
        & CameraCtrl & 13.7 & 0.506 & 0.218 & 47.7 & -- & -- \\
        & MotionCtrl & 13.0 & 0.527 & 0.186 & 49.4 & -- & -- \\
        & CamI2V     & 13.5 & 0.508 & 0.209 & 46.6 & -- & -- \\
        & RealCam    & 13.2 & 0.533 & 0.220 & 52.3 & -- & -- \\
        & Ours       & \textbf{14.9} & \textbf{0.442} & \textbf{0.280} & \textbf{36.4} & \textbf{0.315} & -- \\
        \bottomrule
    \end{tabular}}
    \caption{Quantitative comparison with CamI2V, MotionCtrl, CameraCtrl, and RealCam-I2V on KITTI, Waymo, RealEstate10K, and DL3DV-10K datasets. Inference time is reported only for KITTI.}
    \vspace{3pt}
    \label{tab:quan_results}
\end{table}

\subsection{Training} \label{sec:training}

We train all model components end-to-end. To handle the uncertain velocities, we introduce a variational encoder comprising \(\mathrm{enc_d}\) and \(\mathrm{enc_{vae}}\), that predicts the mean and standard deviation of the latent \(z\), conditioned on the next frame \(I_{t+1}\) and its estimated depth-map \(D_{t+1}\), as well as features of the input frame $I_t$ from \(\mathrm{enc_s}\).
\(\mathrm{enc_d}\) employs skip connections from \(\mathrm{enc_s}\) to capture object motion between frames. Similar to \(\mathrm{enc_s}\), \(\mathrm{enc_d}\) incorporates DINOv2 features extracted from \(I_{t+1}\).
Thus, we have:
\begin{align}
    L_d=\mathrm{enc_d}(I_{t+1}, D_{t+1})   &+ \mathrm{conv_d}(\text{DINOv2}(I_{t+1})) \\
    \mu_{recon}, \sigma^2_{recon} &= \mathrm{enc_{vae}}(L_d) \\
    z_{recon} &\sim \mathcal{N}(\mu_{recon}, \sigma^2_{recon}) \\
    \quad V, A = \mathrm{dec_d}(\mathrm{conv}(L_s) &+ \mathrm{dec_{vae}}(z_{recon}))
\end{align}

Minibatches consist of the input frame \(I_t\), the subsequent frame \(I_{t+1}\), and two future frames \(I_{t+T}\) and \(I_{t+t_r}\) from the same video, where \(\delta t = T\) and \(\delta t = t_r\), with \(t_r \in \{0, \dots, T-1\}\). Given \(I_t\), the model predicts Gaussian parameters, updates Gaussian positions for the respective time intervals, and renders frames and depth-maps at \(\delta t = T\) and \(\delta t = t_r\) using ground-truth camera poses \(\delta\pi_T\) and \(\delta\pi_{t_r}\). All encoder and decoder parameters, as well as the convolutional layers \(\mathrm{conv}\), \(\mathrm{conv_s}\), and \(\mathrm{conv_d}\), are jointly optimized via gradient descent on the loss \(\mathcal{L}\): 
%
\begin{equation}
    I_{t+T}', D_{t+T}' = \mathrm{render}(G_{t+T}, \delta\pi_T) 
\end{equation}
\begin{equation}
    I_{t+t_r}', D_{t+t_r}' = \mathrm{render}(G_{t+t_r}, \delta\pi_{t_r})  
\end{equation}
\begin{equation}
    {\cal L}_{\rm rgbDiff} = {\mathrm{L1}(I_{t+T}-I_t,I_{t+T}' - I_t')}  
\end{equation}
\begin{equation}
\begin{split}
\mathcal{L}_{\mathrm{rgb}} &=
\lambda_1\, \mathrm{LPIPS}(I_{t+T}, I_{t+T}') \\
&\quad + (1-\lambda_1)\,
\mathrm{LPIPS}(I_{t+t_r}, I_{t+t_r}')
\end{split}
\end{equation}
\begin{equation}
\begin{split}
\mathcal{L}_{\mathrm{depth}} &=
\lambda_2\, \mathrm{MRE}(D_{t+T}, D_{t+T}') \\
&\quad + (1-\lambda_2)\,
\mathrm{MRE}(D_{t+t_r}, D_{t+t_r}')
\end{split}
\end{equation}
\begin{equation}
    {\cal L}_{\rm kl} = D_\mathrm{KL}\left( q(z \mid I_{t+1}, I_t) \,\|\, p(z) \right) 
\end{equation}
\begin{equation}
    {\cal L} = \lambda_{\rm rgb} {\cal L}_{\rm rgb} + \lambda_{\rm depth} {\cal L}_{\rm depth} + \lambda_{\rm rgbDiff} {\cal L}_{\rm rgbDiff} + \lambda_{\rm kl} {\cal L}_{\rm kl}
    \label{eq:loss}
\end{equation}
%
where $\lambda_1$, $\lambda_2$, $\lambda_{\rm rgb}$, $\lambda_{\rm depth}$, $\lambda_{\rm rgbDiff}$ and $\lambda_{\rm kl}$ are weighting hyperparameters. When computing ${\cal L}_{\rm depth}$, we clamp values at 10 to mitigate the influence of outlier pixels.

\section{Experiments}

\begin{table}[t]
\centering
\fontsize{9pt}{10pt}\selectfont
\setlength{\tabcolsep}{4pt}
    \centering
    \resizebox{\linewidth}{!}{%
    \begin{tabular}{@{}l c@{\hspace{4pt}}c@{\hspace{4pt}}c@{\hspace{4pt}}c@{\hspace{4pt}}c@{}}
        \toprule
        Model & PSNR $\uparrow$ & LPIPS $\downarrow$ & SSIM $\uparrow$ & FVD $\downarrow$ & Depth $\downarrow$\\
        \midrule
        const.~velocities\hspace{-3pt} & 19.3 & 0.356 & 0.541 & \textbf{30.7} & \textbf{0.308} \\
        with accel. & \textbf{19.4} & \textbf{0.352} & \textbf{0.553} & 30.9 & 0.343 \\
        \bottomrule
    \end{tabular}}
    \caption{Ablation on splat acceleration (Waymo 256$\times$256)}
    \vspace{1pt}
    \label{tab:ablations_acc}
\end{table}
\begin{table}[t]
    \centering
    \resizebox{\linewidth}{!}{%
    \begin{tabular}{@{}l c@{\hspace{4pt}}c@{\hspace{4pt}}c@{\hspace{4pt}}c@{\hspace{4pt}}c@{}}
        \toprule
        Model & \hspace{-4pt}PSNR $\uparrow$ & LPIPS $\downarrow$ & SSIM $\uparrow$ & FVD $\downarrow$ & Depth $\downarrow$\\
        \midrule
        w/o velocities  & 20.2 & 0.319 & 0.572 & 14.2 & 0.210 \\
        det. velocities\hspace{-6pt} & 20.6 & 0.293 & 0.597 & 14.5 & 0.229 \\
        1-Gaussian & 20.1 & 0.325 & 0.578 & 16.1 & 0.210 \\
        w/o DINOv2 & \textbf{21.0} & 0.290 & 0.625 & 14.6 & \textbf{0.192} \\
        Ours & \textbf{21.0} & \textbf{0.288} & \textbf{0.626} & \textbf{14.1} & 0.193 \\
        \bottomrule
    \end{tabular}}
    \caption{Other ablations with constant splat velocities (Waymo 256$\times$832)}
    \vspace{1pt}
    \label{tab:ablations}
\end{table}

\paragraph{Datasets.}
We evaluate our method on four diverse datasets: KITTI~\cite{geiger2013vision} and Waymo Open Perception~\cite{mei2022waymo}, which depict urban driving scenes; DL3DV-10K~\cite{Ling_2024_CVPR}, which spans diverse real-world environments; and RealEstate10K~\cite{Zhou2018StereoMagnification}, which includes indoor and outdoor residential scenes. We train on 60 KITTI sequences and 819 Waymo sequences, evaluating on 3 and 50 sequences, respectively. Waymo sequences have ~198 frames each, while KITTI ranges from 500–5000 frames. For DL3DV-10K, we use 890 scenes for training and 20 for testing. For RealEstate10K, we follow the official splits but use only 5000 training and 100 test scenes, each with 50–200 frames, sampled at a frame rate five times higher than the original.

\begin{figure*}[t]
    \centering

    \begin{minipage}{0.165\textwidth}
        \centering \scriptsize Input
    \end{minipage}%
    \begin{minipage}{0.165\textwidth}
        \centering \scriptsize GT
    \end{minipage}%
    \begin{minipage}{0.165\textwidth}
        \centering \scriptsize w/o velocities (RGB)
    \end{minipage}%
    \begin{minipage}{0.165\textwidth}
        \centering \scriptsize w/o velocities (depth)
    \end{minipage}%
    \begin{minipage}{0.165\textwidth}
        \centering \scriptsize with velocities (RGB)
    \end{minipage}%
    \begin{minipage}{0.165\textwidth}
        \centering \scriptsize with velocities (depth)
    \end{minipage}

    \begin{minipage}{0.165\textwidth}
        \centering
        \includegraphics[width=\linewidth]{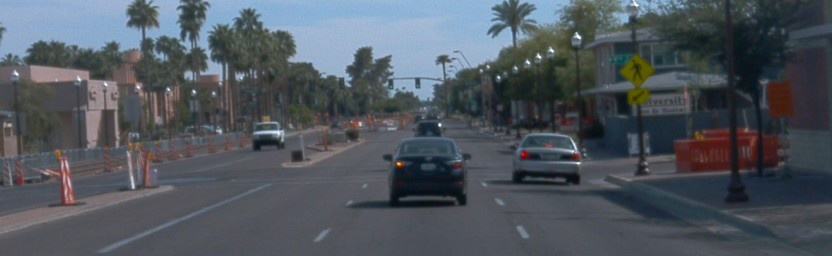}
    \end{minipage}%
    \begin{minipage}{0.165\textwidth}
        \centering
        \includegraphics[width=\linewidth]{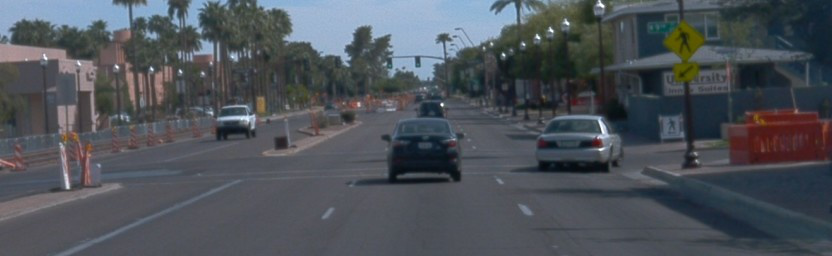}
    \end{minipage}%
    \begin{minipage}{0.165\textwidth}
        \centering
        \includegraphics[width=\linewidth]{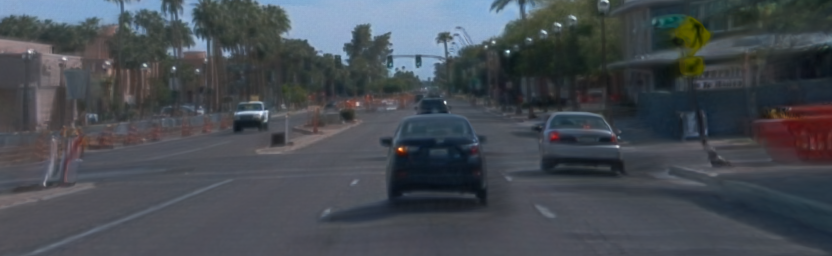}
    \end{minipage}%
    \begin{minipage}{0.165\textwidth}
        \centering
        \includegraphics[width=\linewidth]{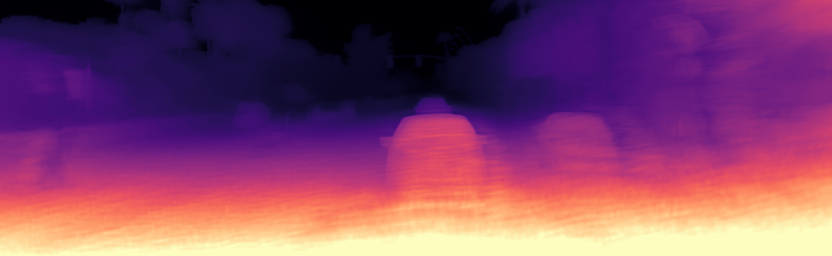}
    \end{minipage}%
    \begin{minipage}{0.165\textwidth}
        \centering
        \includegraphics[width=\linewidth]{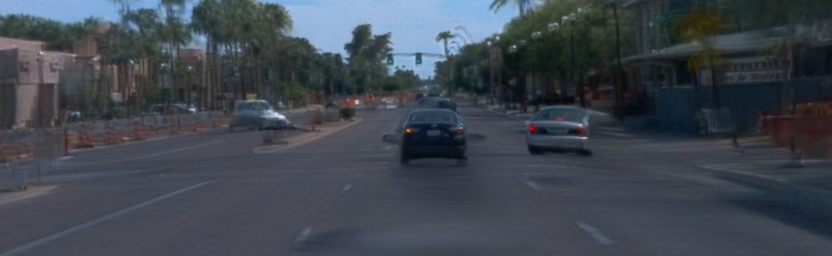}
    \end{minipage}%
    \begin{minipage}{0.165\textwidth}
        \centering
        \includegraphics[width=\linewidth]{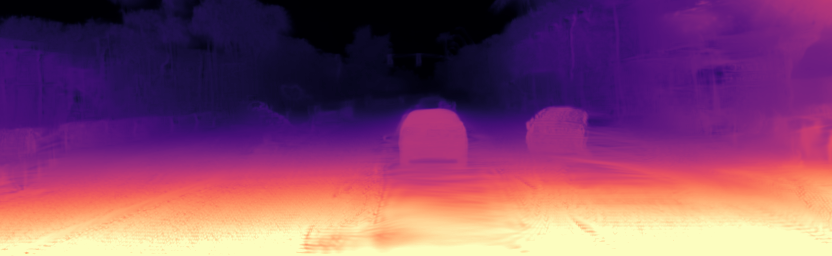}
    \end{minipage}

    \caption{Qualitative ablation results on Waymo: Input and predicted frames and depths at $t+0.9s$. The model with velocities has successfully captured the motion of the black car.}
    \label{fig:abl-static}
\end{figure*}

\begin{figure*}[t]
    \centering
    \tiny


    \begin{minipage}{0.165\textwidth}
        \centering \scriptsize Input
    \end{minipage}%
    \begin{minipage}{0.165\textwidth}
        \centering \scriptsize GT
    \end{minipage}%
    \begin{minipage}{0.165\textwidth}
        \centering \scriptsize det. Velocities (RGB)
    \end{minipage}%
    \begin{minipage}{0.165\textwidth}
        \centering \scriptsize det. Velocities (depth)
    \end{minipage}%
    \begin{minipage}{0.165\textwidth}
        \centering \scriptsize gen. Velocities (RGB)
    \end{minipage}%
    \begin{minipage}{0.165\textwidth}
        \centering \scriptsize gen. Velocities (depth)
    \end{minipage}

    \begin{minipage}{0.165\textwidth}
        \centering
        \includegraphics[width=\linewidth]{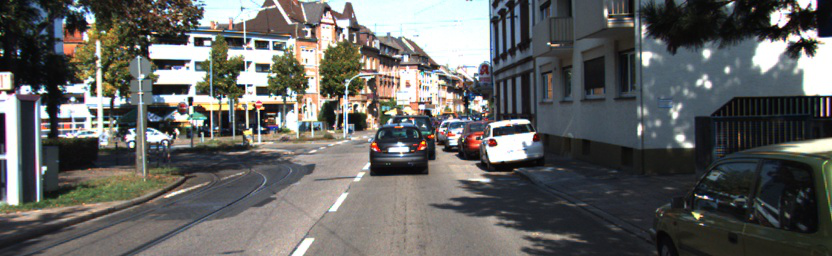}
    \end{minipage}%
    \begin{minipage}{0.165\textwidth}
        \centering
        \includegraphics[width=\linewidth]{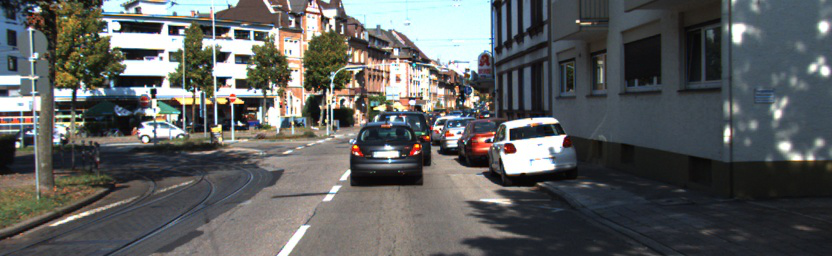}
    \end{minipage}%
    \begin{minipage}{0.165\textwidth}
        \centering
        \includegraphics[width=\linewidth]{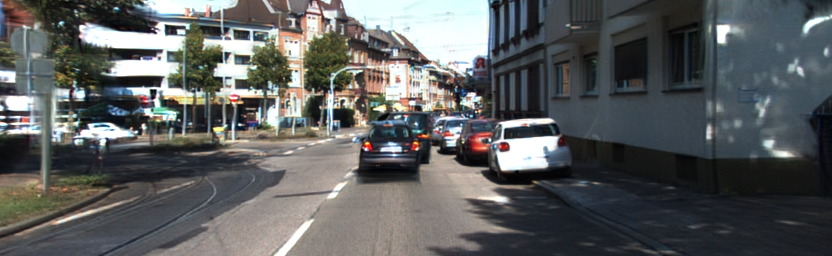}
    \end{minipage}%
    \begin{minipage}{0.165\textwidth}
        \centering
        \includegraphics[width=\linewidth]{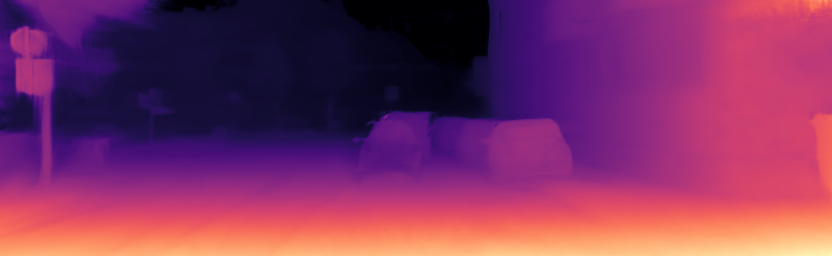}
    \end{minipage}%
    \begin{minipage}{0.165\textwidth}
        \centering
        \includegraphics[width=\linewidth]{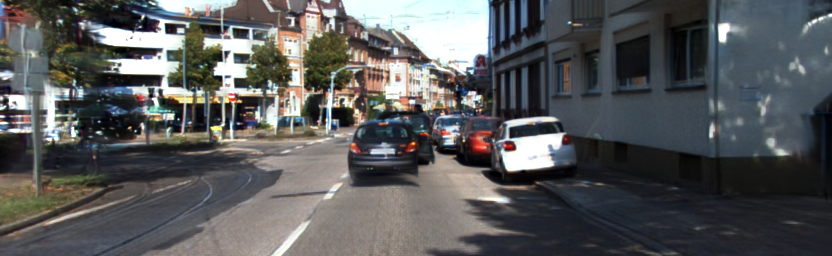}
    \end{minipage}%
    \begin{minipage}{0.165\textwidth}
        \centering
        \includegraphics[width=\linewidth]{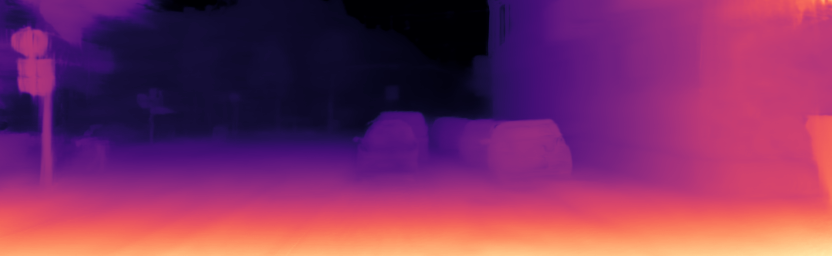}
    \end{minipage}

    \begin{minipage}{0.165\textwidth}
        \centering \scriptsize Input
    \end{minipage}%
    \begin{minipage}{0.165\textwidth}
        \centering \scriptsize GT
    \end{minipage}%
    \begin{minipage}{0.165\textwidth}
        \centering \scriptsize w/o DINOv2
    \end{minipage}%
    \begin{minipage}{0.165\textwidth}
        \centering \scriptsize w/o DINOv2 (depth)
    \end{minipage}%
    \begin{minipage}{0.165\textwidth}
        \centering \scriptsize with DINOv2 (RGB)
    \end{minipage}%
    \begin{minipage}{0.165\textwidth}
        \centering \scriptsize with DINOv2 (depth)
    \end{minipage}

    \begin{minipage}{0.165\textwidth}
        \centering
        \includegraphics[width=\linewidth]{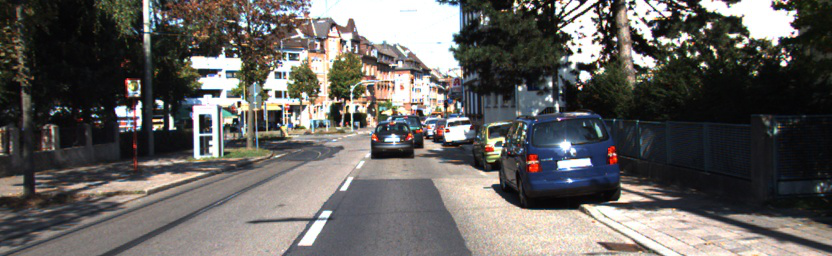}
    \end{minipage}%
    \begin{minipage}{0.165\textwidth}
        \centering
        \includegraphics[width=\linewidth]{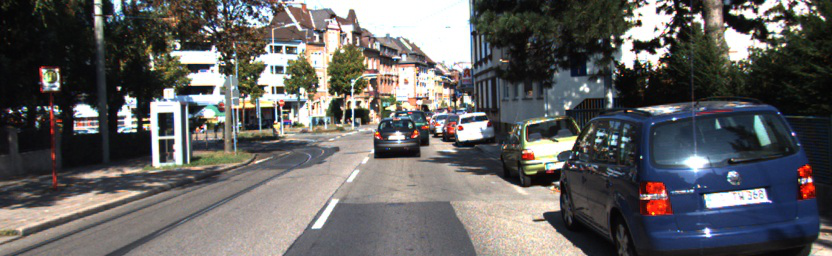}
    \end{minipage}%
    \begin{minipage}{0.165\textwidth}
        \centering
        \includegraphics[width=\linewidth]{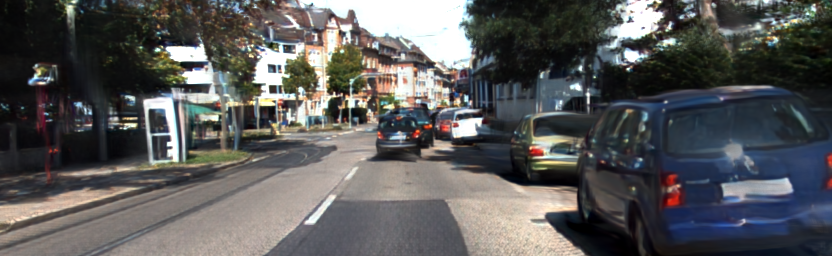}
    \end{minipage}%
    \begin{minipage}{0.165\textwidth}
        \centering
        \includegraphics[width=\linewidth]{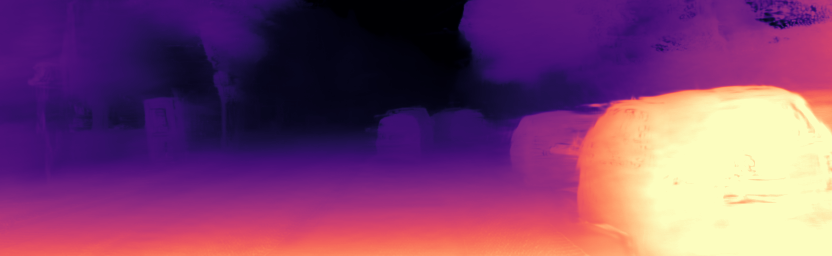}
    \end{minipage}%
    \begin{minipage}{0.165\textwidth}
        \centering
        \includegraphics[width=\linewidth]{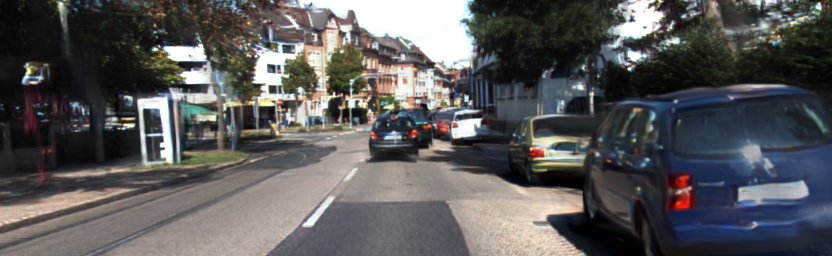}
    \end{minipage}%
    \begin{minipage}{0.165\textwidth}
        \centering
        \includegraphics[width=\linewidth]
        {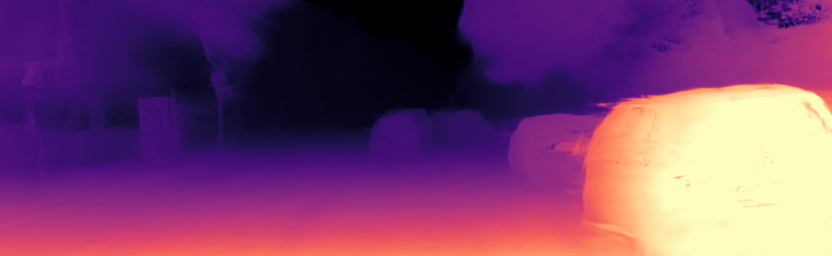}
    \end{minipage}

    \begin{minipage}{0.165\textwidth}
        \centering \scriptsize Input
    \end{minipage}%
    \begin{minipage}{0.165\textwidth}
        \centering \scriptsize GT
    \end{minipage}%
    \begin{minipage}{0.165\textwidth}
        \centering \scriptsize 1-Gaussian (RGB)
    \end{minipage}%
    \begin{minipage}{0.165\textwidth}
        \centering \scriptsize 1-Gaussian (depth)
    \end{minipage}%
    \begin{minipage}{0.165\textwidth}
        \centering \scriptsize 5-Gaussians (RGB)
    \end{minipage}%
    \begin{minipage}{0.165\textwidth}
        \centering \scriptsize 5-Gaussians (depth)
    \end{minipage}

    \begin{minipage}{0.165\textwidth}
        \centering
        \includegraphics[width=\linewidth]{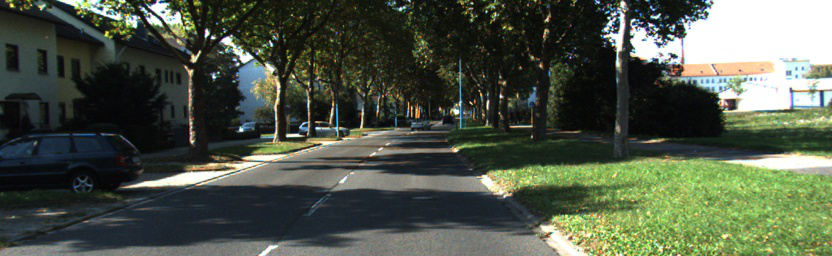}
    \end{minipage}%
    \begin{minipage}{0.165\textwidth}
        \centering
        \includegraphics[width=\linewidth]{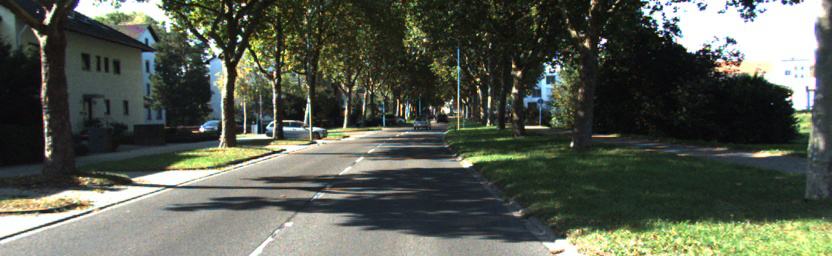}
    \end{minipage}%
    \begin{minipage}{0.165\textwidth}
        \centering
        \includegraphics[width=\linewidth]{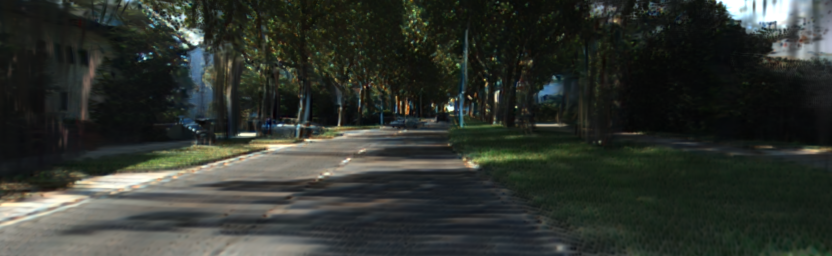}
    \end{minipage}%
    \begin{minipage}{0.165\textwidth}
        \centering
        \includegraphics[width=\linewidth]{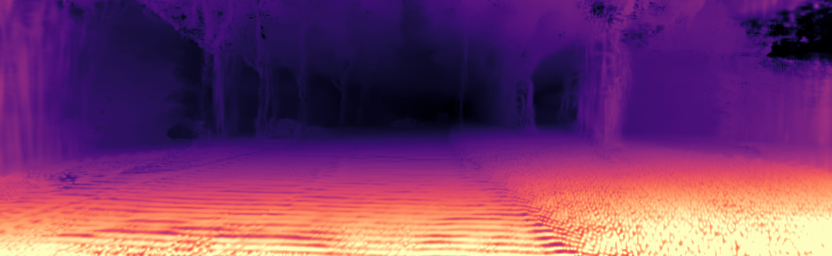}
    \end{minipage}%
    \begin{minipage}{0.165\textwidth}
        \centering
        \includegraphics[width=\linewidth]{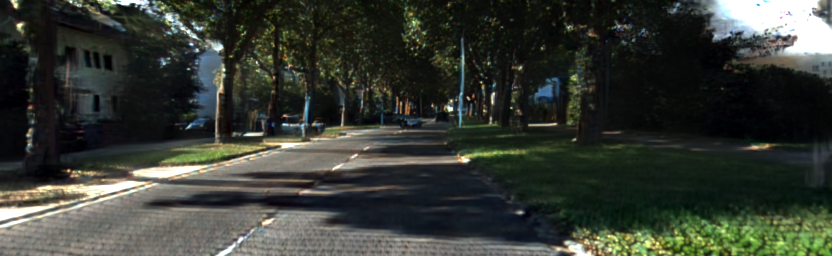}
    \end{minipage}%
    \begin{minipage}{0.165\textwidth}
        \centering
        \includegraphics[width=\linewidth]{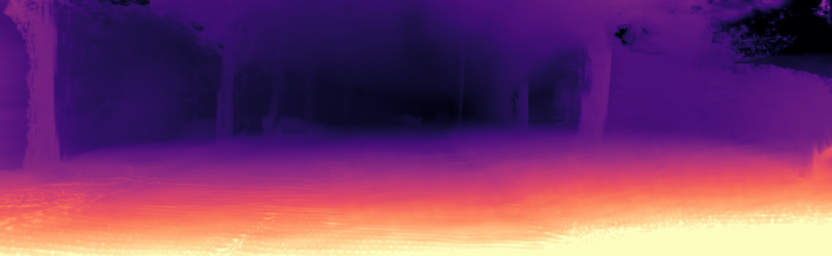}
    \end{minipage}
    \caption{
Qualitative ablation results on KITTI, showing input and predicted frames and depths at $t+0.9s$. \textbf{Top:} with deterministic velocities and generative velocities. Motion of the black car is more realistic in the gen. velocities version. 
\textbf{Middle:} without DINOv2 and with DINOv2. The appearance of the black car is more realistic with DINOv2.
\textbf{Bottom:} with 1 Gaussian per pixel and 5 Gaussians per pixel. Around nearby trees and the road are blurry in the 1-Gaussian version.
    }
    \label{fig:abl-dino-and-1-splat}
\end{figure*}

\paragraph{Metrics.}
We evaluate the generated future frames using the average peak signal-to-noise ratio (PSNR), structural similarity index (SSIM), and learned perceptual image patch similarity (LPIPS), while Fréchet
Video Distance (FVD) measures distributional
similarity between generated and real video clips, and the average mean relative depth error measures the quality of 4D representations.
These metrics quantify the similarity between generated and original videos, with high scores requiring both accurate camera motion and realistic object motion. Note that we do not use motion fidelity metrics such as RotErr, TransErr, and CamMC \cite{li2025realcam,zheng2024cami2v}, since these are limited to static scenes and do not give reliable results on fast forward-moving cameras, as in KITTI~\cite{Geiger2012CVPR} and Waymo~\cite{mei2022waymo}.

\paragraph{Baselines.}
We compare our method against four recent camera-guided image-to-video models: RealCam-I2V~\cite{li2025realcam}, CamI2V~\cite{zheng2024cami2v}, CameraCtrl~\cite{he2024cameractrl}, and MotionCtrl~\cite{wang2023motionctrl}. CameraCtrl leverages Plücker embeddings of camera poses for text-to-video generation, MotionCtrl integrates camera pose with temporal transformers, and CamI2V employs epipolar attention for image-to-video synthesis.  
RealCam-I2V~\cite{li2025realcam} reconstructs and conditions on point clouds for 3D consistency, with dynamic content modeled by a video diffusion model.

\paragraph{Implementation details.}
We adopt the Song-UNet architecture for \(\mathrm{enc_s}\), \(\mathrm{enc_d}\) and \(\mathrm{dec_s}\)~\cite{song21ddim,Szymanowicz2023-lx}. \(\mathrm{dec_d}\) is a 5-stage upsampling decoder using Song-UNet blocks, producing 6-channel outputs at the input resolution. For \(\mathrm{vae}\), we use a standard convolutional encoder-decoder with Group Normalization and LeakyReLU activations; the encoder downsamples the output of \(\mathrm{enc_d}\) through four stages to a compact latent code, which the decoder symmetrically upsamples.
We first train \(\mathrm{enc_s}\), \(\mathrm{dec_s}\), and \(\mathrm{conv_s}\) without dynamics, then fine-tune them alongside \(\mathrm{enc_d}\), \(\mathrm{dec_d}\), \(\mathrm{vae}\), \(\mathrm{conv_d}\), and \(\mathrm{conv}\) trained from scratch. We set \(K=5\) and predicts 1.6s \(256 \times 256\) videos. Ablations with constant velocities use 1s \(832 \times 256\) videos.
Further details on hyperparameters are provided in the Supplementary Material.

\subsection{Comparison against State-of-the-art Camera-controlled Video Synthesis Methods} \label{sec:eval}

The quantitative results in Table~\ref{tab:quan_results} show that our method outperforms all baselines in four datasets and metrics - including FVD, PSNR, LPIPS, and SSIM - while also achieving faster inference. In the Waymo data set, which features multiple dynamic objects per scene, our approach achieves an FVD of 30.9 (vs.~MotionCtrl’s 43.3), a PSNR of 19.4 (vs.~ RealCam-I2V's 17.0), and an SSIM of 0.553 (vs.~RealCam-I2V’s 0.424), despite not using any of these metrics during training. Similarly, on DL3DV-10K, which features diverse scenes, varying motion patterns, and complex camera trajectories, our method achieves an FVD of 36.4, outperforming all baselines. The low Fréchet Video Distance indicates that our generated videos are not only temporally consistent but also visually realistic.
The qualitative results in Figure~\ref{fig:qual_results} show that our method preserves temporal and 3D consistency while faithfully following the details of the image. Its underlying 4D representation ensures accurate camera motion. 

\subsection{Ablation Study}\label{sec:ablation}

\paragraph{Gaussian velocities and acceleration.} 
To evaluate the importance of incorporating linear and angular accelerations, we conduct experiments both with full acceleration modeling and with a simplified `constant-velocity' baseline. Table \ref{tab:ablations_acc} shows that the `with-acceleration' model achieves better frame-level visual quality (PSNR, SSIM, LPIPS), whereas the simpler `constant-velocity' variant yields lower FVD and more accurate depth.
We also conduct an experiment that removes motions entirely, reducing our 4D scene to 3D.
Table~\ref{tab:ablations} shows that `ours' outperforms `w/o velocities' across all metrics, demonstrating its effectiveness in capturing scene dynamics. 
Figure~\ref{fig:abl-static} shows that our method successfully predicts the motion of the vehicle in the scene, while the static version naturally cannot model this movement. 
On the other hand, motion parameters of objects, and thus our Gaussian splats, are inherently uncertain given a single image. Therefore, they should be sampled from a learned conditional distribution, rather than being regressed as a single deterministic estimate. Assuming constant acceleration, we compare the performance of our generative approach against a deterministic variant that directly predicts point estimates of velocity, conditioned on \(L_s\), setting.
$
V = \mathrm{dec_d}(\mathrm{conv}(L_s))
$.
Table~\ref{tab:ablations} shows that `ours' outperforms `det.~velocities' across all evaluation metrics. Additionally, as shown in Figure~\ref{fig:abl-dino-and-1-splat} (top, black car and its depths), the model with gen. Velocities have shown sharper and more realistic object motion than the model with det. Velocities.

\paragraph{Multiple splats per pixel.}
We argue that predicting multiple Gaussians per pixel is crucial for quality, to avoid gaps in the scene. As shown in Figure~\ref{fig:abl-dino-and-1-splat} (bottom), the 1-Gaussian model leaves noticeable voids in later frames (particularly around nearby trees and the road), compared to the 5-Gaussian model. Additionally, Table~\ref{tab:ablations} demonstrates that `ours', which uses five Gaussian splats per pixel, outperforms the `1-Gaussian' model across all evaluation metrics.

\paragraph{DINOv2 features.}
While our model reconstructs pixel-aligned Gaussians via a pixel-to-pixel U-Net, we find that incorporating DINOv2 features~\cite{dino2023} enriches the encoder with semantic context, leading to better depth refinement and xy-offset estimation. Figure~\ref{fig:abl-dino-and-1-splat} (middle) shows that removing DINOv2 degrades quality, particularly in moving objects. For example, the black car appears noticeably sharper in future frames when DINOv2 is used. Additionally, Table~\ref{tab:ablations} shows that `ours' outperforms `w/o DINOv2' across FVD, LPIPS, and SSIM.

\section{Conclusion}

We have presented Pixel-to-4D, a method for camera-controlled single-image to video generation with an intermediate 4D representation.
Our approach achieves precise camera motion and temporal consistency for free by using our 4D representation to render future frames. 
Furthermore, unlike existing single-image-to-4D reconstruction methods, our approach requires just one forward pass without using any test-time optimization or diffusion priors, which greatly reduces inference time.
We have shown that Pixel-to-4D achieves state-of-the-art results on camera-controlled video prediction on four real-world datasets.

{
    \small
    \bibliographystyle{ieeetr}
    \bibliography{main}
}


\end{document}